\documentclass[sigconf, nonacm=true]{acmart}
\usepackage[utf8]{inputenc}
\usepackage[Export]{adjustbox} 

\usepackage{amsmath}

\DeclareMathOperator*{\argmin}{argmin}

\AtBeginDocument{%
  \providecommand\BibTeX{{%
    \normalfont B\kern-0.5em{\scshape i\kern-0.25em b}\kern-0.8em\TeX}}}

\setcopyright{acmcopyright}
\copyrightyear{2023}
\acmYear{2023}
\acmDOI{XXXXXXX.XXXXXXX}

\acmConference[Conference acronym 'XX]{Make sure to enter the correct
  conference title from your rights confirmation emai}{June 03--05,
  2018}{Woodstock, NY}
\acmPrice{15.00}
\acmISBN{978-1-4503-XXXX-X/18/06}

\begin{document}

\title{Minimising changes to audit when updating decision trees}
\author{Anj Simmons}
\affiliation{%
  \institution{Hashtag AI}
  \city{Melbourne}
  \country{Australia}
}
\email{anj@simmons.ai}

\author{Scott Barnett}
\affiliation{%
  \institution{Deakin University}
  \city{Geelong}
  \country{Australia}
}
\email{scott.barnett@deakin.edu.au}

\author{Anupam Chaudhuri}
\affiliation{%
  \institution{Deakin University}
  \city{Geelong}
  \country{Australia}
}
\email{anupam.chaudhuri@deakin.edu.au}

\author{Sankhya Singh}
\affiliation{%
  \institution{Deakin University}
  \city{Geelong}
  \country{Australia}
}
\email{sankhya.singh@deakin.edu.au}

\author{Shangeetha Sivasothy}
\affiliation{%
  \institution{Deakin University}
  \city{Geelong}
  \country{Australia}
}
\email{s.sivasothy@deakin.edu.au}

\begin{abstract}
Interpretable models are important, but what happens when the model is updated on new training data? We propose an algorithm for updating a decision tree while minimising the number of changes to the tree that a human would need to audit. We achieve this via a greedy approach that incorporates the number of changes to the tree as part of the objective function. We compare our algorithm to existing methods and show that it sits in a sweet spot between final accuracy and number of changes to audit.

\end{abstract}



\keywords{Decision trees, pruning, interpretability}

\maketitle

\section{Introduction}


Consider the case of a machine learning model deployed in a safety critical environment or used to make important decisions about humans. For example, a model to decide which patients are prioritised for treatment. Blackbox machine learning models are unsuitable for this task, as they behave unpredictably when used on data that differs to their training data, and suffer from bias and fairness issues. Interpretable machine learning models are preferable \cite{Rudin2019}, as they provide the ability for a human expert to audit the rules the model has learned to ensure they are safe and fair. For example, decision trees are an interpretable model that can be easily understood by clinicians, thus allowing clinicians to provide expert approval of the model prior to use on patients and to reject the model if it contains rules that could endanger patients or violate ethical practices.

As new data becomes available, it would be desirable to update the model. However, as retraining the model results in a completely new ruleset, and auditing a machine learning model is a time-consuming process, this results in outdated models to continue being used until sufficient resources are available to retrain and audit a new model. In this paper, we propose a modified decision tree learning algorithm that minimises structural changes during retraining, thus reducing the cost and time to audit the updated model. An implementation of our algorithm is provided online\footnote{\url{https://github.com/a2i2/tree_diff/}}. 


\section{Objective Function}


Consider a streaming version of the supervised learning setting where data is available in batches $\{b_0, b_1, ..., b_n\}$. At time point $t$, a batch $B_t$ of $(x, y)$ pairs is used as training data where $x$ is the features and $y$ is the associated label. A classifier that produces a ruleset $r_{t}$ (i.e. a decision tree) is updated by incrementally training on each batch. The full set of data available at time $t$ is $D_t = \bigcup_{i \in 0..t} {B}_i$.

At each step, a human then reviews the ruleset and approves or rejects the updates. To facilitate this review, in addition to minimising misclassifications, $f(r_t, D_t)$, we need to consider the complexity of the ruleset, $c(r_t)$, and the number of changes made to the ruleset, $\Delta(r_{t-1}, r_t)$. This is incorporated into a loss function, $L(r_{t-1}, r_{t}, D_t)$ shown in \autoref{eq:loss}.

\begin{equation} \label{eq:loss}
L(r_{t-1}, r_{t}, D_t) = f(r_{t}, D_t) + \alpha c(r_{t}) + \beta \Delta(r_{t-1}, r_t)
\end{equation}

For the purposes of this paper, we focus on rulesets that are binary decision trees. The complexity, $c(r_{t})$, is measured as the total number of nodes (decision nodes and leaf nodes) in the tree. The number of changes made to the tree $\Delta(r_{t-1}, r_t)$ is measured as the number of new nodes introduced in $r_t$ that were not present in $r_{t-1}$ (altering the variable or threshold of a decision node is treated as discarding that node and all its descendants). The number of misclassifications, $f(r_t, D_t)$ is the total number of points in the dataset that are not correctly classified by a leaf node. The number of misclassifications is measured on the training dataset rather than a validation dataset (which is not available when creating the tree), hence should be used in combination with $c(r_{t})$ to penalise complex trees that overfit the data.


$\alpha$ controls the penalty applied to complex trees. A small value of $\alpha$ will result in deep trees that overfit the data. A large value of $\alpha$ will result in a high degree of pruning, which simplifies the tree, but may not capture nuanced patterns in the data. $\beta$ controls the penalty applied to changes to the tree. A small value of $\beta$ will completely discard the original tree if a better tree can be found. In contrast, a high value of $\beta$ will tend to preserve the original tree structure.

\begin{figure*}[t]
    \centering
    \includegraphics[width=0.49\linewidth]{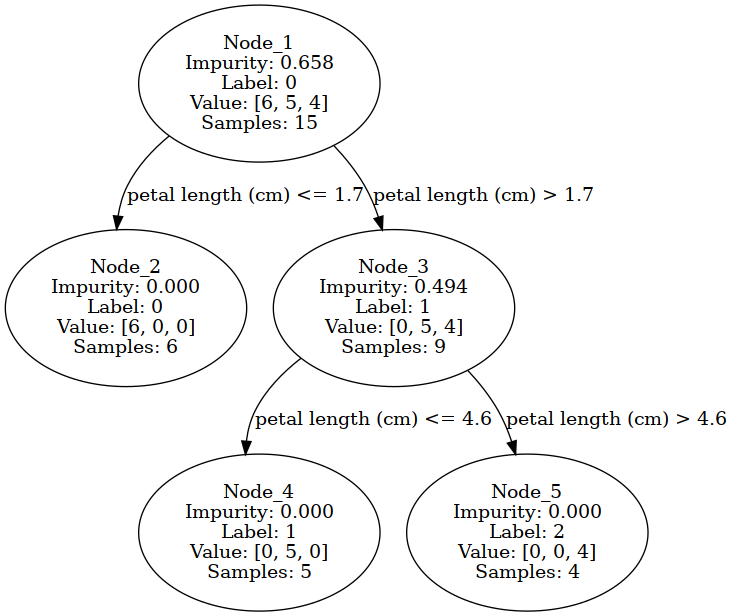}
    \includegraphics[width=0.49\linewidth]{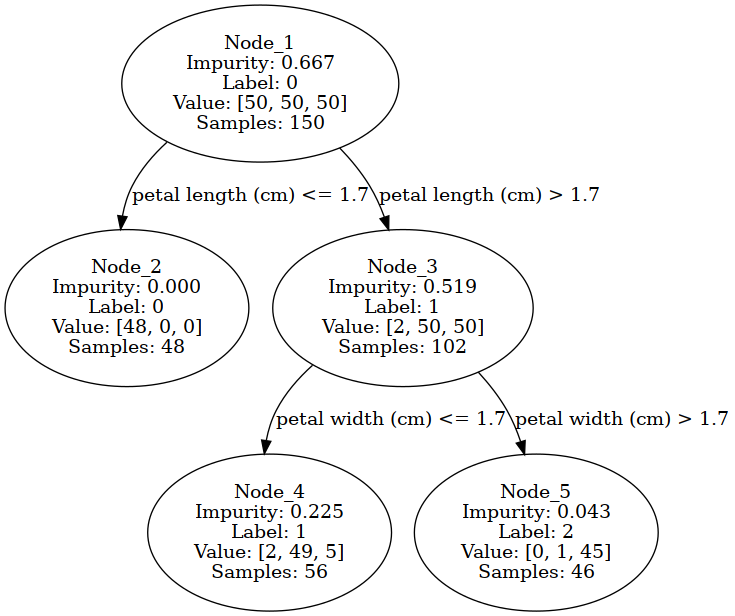}
    \caption{Initial decision tree at t=0 (left), and updated tree at t=1 (right) produced by our algorithm. Note how our algorithm `keeps' Node 1 and Node 2 to minimise changes to audit and `regrows' Node 3 for improved accuracy.}
    \label{fig:example}
\end{figure*}

\section{Keep-Regrow Algorithm}

In this section, we describe the Keep-Regrow algorithm.

A tree is represented as a tuple of (variable, threshold, left subtree, right subtree), as shown in \autoref{eq:repr}. In the case of a leaf node, it is represented as the class label for that leaf.

\begin{align}
\begin{split}\label{eq:repr}
r_t = \begin{cases} 
        (V(r_t), T(r_t), S_L(r_t), S_R(r_t)) & N(r_t) \\
        C(r_t) & \text{otherwise}
      \end{cases}
\end{split}
\end{align}

Where $V$ is the top level split variable, $T$ is the threshold, $S_L$ is the ruleset for the left subtree (less than or equal to threshold), and $S_R$ is the right subtree (greater than threshold). $N(r_t)$ is true if $r_t$ is a split node, and false if $r_t$ is a leaf. $C(r_t)$ is the class to be output for leaf $r_t$.

To support an efficient solution to optimising \autoref{eq:loss}, we note that each branch of the tree can be independently optimised, as per \autoref{eq:recurse}. This allows us to take a greedy approach.

\begin{align}
\begin{split}\label{eq:recurse}
L(r_{t-1}, r_t, D_t) ={}& L(S_L(r_{t-1}), S_L(r_{t}), D_L(V(r_t), T(r_t), D_t)) \\
& + L(S_R(r_{t-1}), S_R(r_{t}), D_R(V(r_t), T(r_t), D_t)) \\
& + \alpha \\
& + \begin{cases} 
      \beta & V(r_t) = V(r_{t-1})\, \text{and}\, T(r_t) = T(r_{t-1}) \\
      0 & \text{otherwise}
   \end{cases}
\end{split}
\end{align}

Where $D_L(V(r_t), T(r_t), D_t) = \{(x, y) \in D_t: x[V(r_t)] \leq T(r_t)\} )$ is data that applies to the left subtree, and $D_R(V(r_t), T(r_t), D_t) = \{(x, y) \in D_t: x[V(r_t)] > T(r_t)\} )$ is data that applies to the right subtree. The complexity penalty $\alpha$ is applied to all nodes, whereas the change penalty $\beta$ is only applied if the node variable or threshold differs to the previous tree.

In the case that $r_t$ is a leaf node, the change penalty only applies if the class label changes, as shown in \autoref{eq:leaf}.

\begin{align}
\begin{split}\label{eq:leaf}
L(r_{t-1}, r_t, D_t) ={}& \alpha + \begin{cases} 
      \beta & C(r_t) = C(r_{t-1}) \\
      0 & \text{otherwise}
   \end{cases}
\end{split}
\end{align}

At each node, there are two options. 1) \textit{Keep} the current split, or 2) \textit{Regrow} the tree using a different variable or threshold at that split node. We evaluate the loss from each option, then take the best option, $O(r_{t-1}, D_t)$, that minimises the loss function, as per \autoref{eq:opt}.

\begin{align}
\begin{split}\label{eq:opt}
O(r_{t-1}, D_t) = \argmin_{r\in\{K(r_{t-1}, D_t), R(D_t)\}} L(r_{t-1}, r, D_t)
\end{split}
\end{align}

Where $K$ keeps the split node and $R$ regrows the node with a different split variable or threshold.

\subsection{Keep}

For the keep option, shown in \autoref{eq:keep}, the variable and threshold remain unchanged. However, we optimise the left and right subtrees.

\begin{align}
\begin{split}\label{eq:keep}
K(r_{t-1}, D_t) = (&V(r_{t-1}), T(r_{t-1}), \\
&O(S_L(r_{t-1}), D_L(V(r_{t-1}), T(r_{t-1}), D_t)), \\
&O(S_R(r_{t-1}), D_R(V(r_{t-1}), T(r_{t-1}), D_t)))
\end{split}
\end{align}

If $r_{t-1}$ is a leaf node, the keep option preserves the class label (modifying the class label would introduce a change penalty, and hence falls under the regrow option instead).

\subsection{Regrow}

For the regrow option, shown in \autoref{eq:regrow}, we regrow the tree using a standard tree regrowing algorithm (e.g. CART), then prune the tree to account for the complexity and change penalties. As regrowing discards the previous tree node (i.e. change penalty always applies), the previous tree, $r_{t-1}$ is not an input to this function.

\begin{align}
\begin{split}\label{eq:regrow}
R(D_t) = P(G(D_t), D_t)
\end{split}
\end{align}

Where $P$ is the pruning function, and $G$ is a standard tree growing algorithm. Any node, including leaf nodes, can be regrown. When leaf nodes are regrown, this results in producing a deeper tree. Note that pruning incorporates the possibility of reducing the tree to a single leaf node.

\begin{table*}[h!]
\centering
\begin{tabular}{|l|l|p{6cm}|c|c|}
\hline
\textbf{Item} & \textbf{Dataset Link} & \textbf{Objective} & \parbox{2cm}{\centering \textbf{Predictive\\Features}} & \parbox{2cm}{\centering \textbf{Classification\\Type}} \\ \hline
Skin & \href{https://archive.ics.uci.edu/dataset/229/skin+segmentation}{/dataset/229/skin+segmentation} & Classify pixel as skin or nonskin based on B, G, R color space. & 3 & Binary \\ \hline
HIGGS & \href{https://archive.ics.uci.edu/dataset/280/higgs}{/dataset/280/higgs} & Classify as either a signal process that produces Higgs bosons or background process that does not. & 28 & Binary \\ \hline
SUSY & \href{https://archive.ics.uci.edu/dataset/279/susy}{/dataset/279/susy} & Classify as signal process which produces supersymmetric particles or a background process which does not. & 18 & Binary \\ \hline
HEPMASS & \href{https://archive.ics.uci.edu/dataset/347/hepmass}{/dataset/347/hepmass} & Classify as particle-producing signal process or a background process. & 28 & Binary  \\ \hline
Poker & \href{https://archive.ics.uci.edu/dataset/158/poker+hand}{/dataset/158/poker+hand} & Classify the type of poker hand. & 10 & Multi-class \\ \hline
Cover & \href{https://archive.ics.uci.edu/dataset/31/covertype}{/dataset/31/covertype} & Classify the type of forest cover. & 54 & Multi-class \\ \hline
\end{tabular}
\caption{Datasets used for evaluation. Base URL: \texttt{https://archive.ics.uci.edu}.}
\label{tab:datasets}
\end{table*}

\subsection{Prune}

Pruning is applied after growing a subtree, $r_g$. During the pruning stage, shown in \autoref{eq:prune}, at each split node there are two options. 1) Preserve the \textit{split} node, or 2) \textit{Terminate} the node (convert it to a leaf node). The option that minimises the loss function is selected (as during regrowing the previous tree is irrelevant, i.e. never preserved, we pass $r_{t-1} = \emptyset$ in the loss function). In the case of a leaf node, only the terminate option is applicable.

\begin{align}
\begin{split}\label{eq:prune}
P(r_g, D_t) = \argmin_{r\in\{S(r_g, D_t), T(r_g, D_t)\}} L(\emptyset, r, D_t)
\end{split}
\end{align}

Where $S$ preserves the split node, and $T$ terminates the node.

\subsubsection{Preserve Split}

If the split node is preserved, we recursively prune the left and right subtrees, as shown in \autoref{eq:split}.

\begin{align}
\begin{split}\label{eq:split}
S(r_g, D_t) = (&V(r_g), T(r_g), \\
&P(S_L(r_g), D_L(V(r_g), T(r_g), D_t)), \\
&P(S_R(r_g), D_R(V(r_g), T(r_g), D_t)))
\end{split}
\end{align}

\subsubsection{Terminate}

If the node is terminated, we set it to a leaf node with the best (most frequent) class label, $c$.

\begin{align}
\begin{split}\label{eq:terminate}
T(r_g, D_t) = \argmin_c L(\emptyset, c, D_t)
\end{split}
\end{align}

\section{Example}

We demonstrate the algorithm on the iris dataset. At t=0, we have a random sample of one tenth of the iris dataset. At t=1 we have access to the full iris dataset. \autoref{fig:example} left shows a tree grown at t=0 produced using a CART tree growing algorithm. This is the initial model, which a human can review and make changes to as necessary. \autoref{fig:example} right shows the tree at t=1 produced using the Keep-Regrow algorithm with $\alpha=1$ and $\beta=1$. Note that nodes 1 and 2 of the previous tree have been preserved to make auditing the new tree as simple as possible. Node 3 has been regrown, resulting in a different variable and threshold condition. Nodes 4 and 5 are also considered to be modified as they descend from a different condition to the previous tree. The algorithm determined that the overall complexity, $\alpha{}c(r_1)$, and penalty of making these changes, $\beta\Delta(r_0, r_1)$, was sufficiently offset by the reduction of misclassifications on the full dataset, $f(r_1, D_1)$, to justify the updated tree.






\section{Datasets}

We evaluate the algorithm on large datasets. This was necessary, as the difference in accuracy between algorithms is often a fraction of a percent. By using a test set of 100,000 datapoints, we can estimate accuracy of each run to a 95\% confidence interval of ±0.3\%.

The list of datasets selected for evaluation are shown in \autoref{tab:datasets}. These include a combination of simple binary classification tasks (Skin), Physics datasets (HIGGS, SUSY, HEPMASS), and multi-class classification (Poker, Cover). We randomly sample up to 10 batches of 1,000 training points for use as the training set, and randomly sample 100,000 datapoints from the remaining data as the test set. 

\section{Evaluation}

We re-run each variation of each algorithm 12 times on (up to 10)
batches of 1,000 training points to explore variability arising from the
training data. The number of training points is small compared to the size of the test set, as we are interested in studying the ability of the algorithm to construct good trees given a small dataset, but need to measure the accuracy of these trees to a high degree of precision.

Our evaluation measures the number of nodes (as an approximate measure of tree complexity), accuracy on the held out test set, and similarity of the updated tree trained on the latest batch of data to the tree trained on the previous batches. To permit a fair comparison with other algorithms, we use a pre-existing measure of similarity that treats minor modifications to conditions as a partial match \cite{perner2013compare} rather than the more stringent goal optimised for by our algorithm where any modification to a condition is treated as changing the node and all its descendants.




%
%
%
%

    \hypertarget{exploration-of-alpha-parameter}{%
\subsection{Exploration of Alpha
Parameter}\label{exploration-of-alpha-parameter}}

The alpha parameter controls the penalty for complex trees (measured in
terms of the number of nodes). This can be thought of as a form of
pruning (which is already well studied), but we need to select a value
of alpha so we can proceed with the rest of the study.

The visualisations in this section explore the tradeoff between accuracy
(on the test set) and number of nodes for different values of alpha. Ths
small points show the results for each individual run, and the large
points (outlined in black) show the average estimate across 12 runs for
a particular variant of the algorithm. This is just for the first batch
of data, so is just growing a traditional decision tree (using sklearn)
then performing pruning (with the level controlled by alpha). The
keep-regrow algorithm only comes into play on the second batch of data
once we have an initial decision tree.

On the Skin dataset (consisting of just three dimensions) more nodes
allows a better fit, but on other more complex datasets such as Higgs,
we can see more nodes also leads to overfitting which lowers the
accuracy on the test set.

Based on the results, we suggest a default value of alpha=5 to minimise
the complexity of the tree and prevent overfitting.

    \begin{center}
    \adjustimage{max size={\linewidth}{0.9\paperheight}}{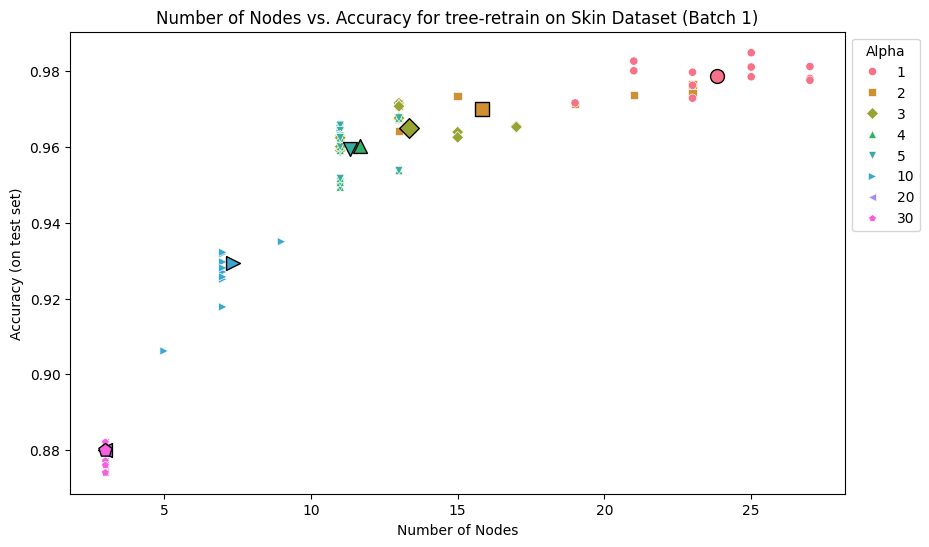}
    \end{center}
    { \hspace*{\fill} \\}
    
    \begin{center}
    \adjustimage{max size={\linewidth}{0.9\paperheight}}{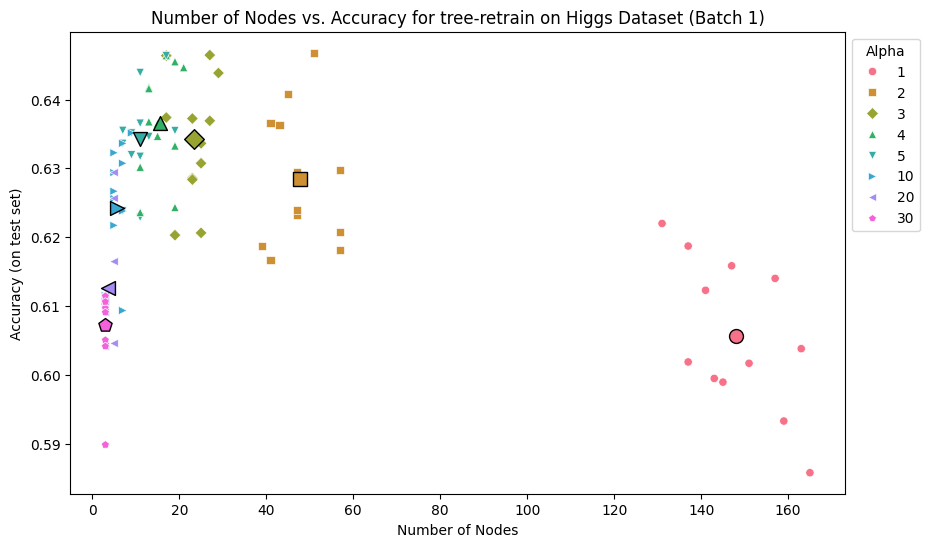}
    \end{center}
    { \hspace*{\fill} \\}
    
    \begin{center}
    \adjustimage{max size={\linewidth}{0.9\paperheight}}{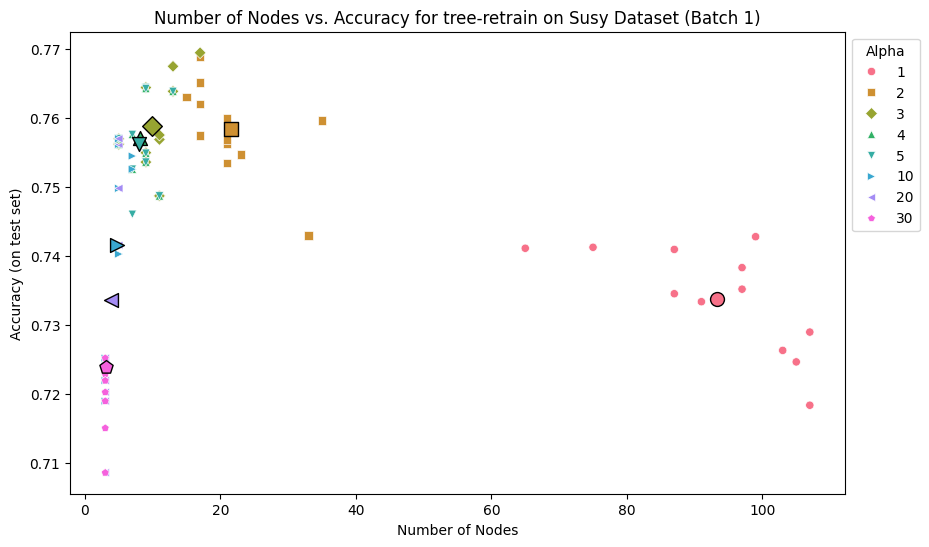}
    \end{center}
    { \hspace*{\fill} \\}
    
    \begin{center}
    \adjustimage{max size={\linewidth}{0.9\paperheight}}{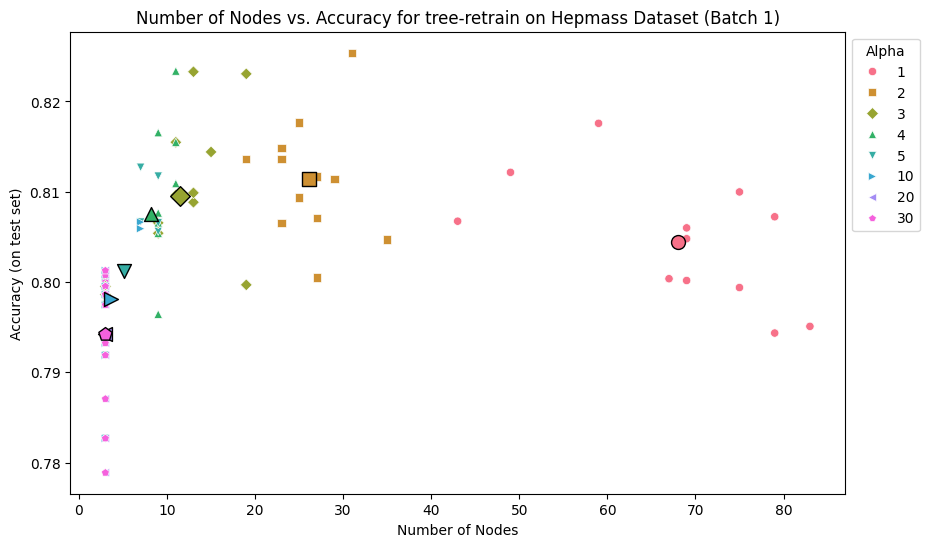}
    \end{center}
    { \hspace*{\fill} \\}
    
    \begin{center}
    \adjustimage{max size={\linewidth}{0.9\paperheight}}{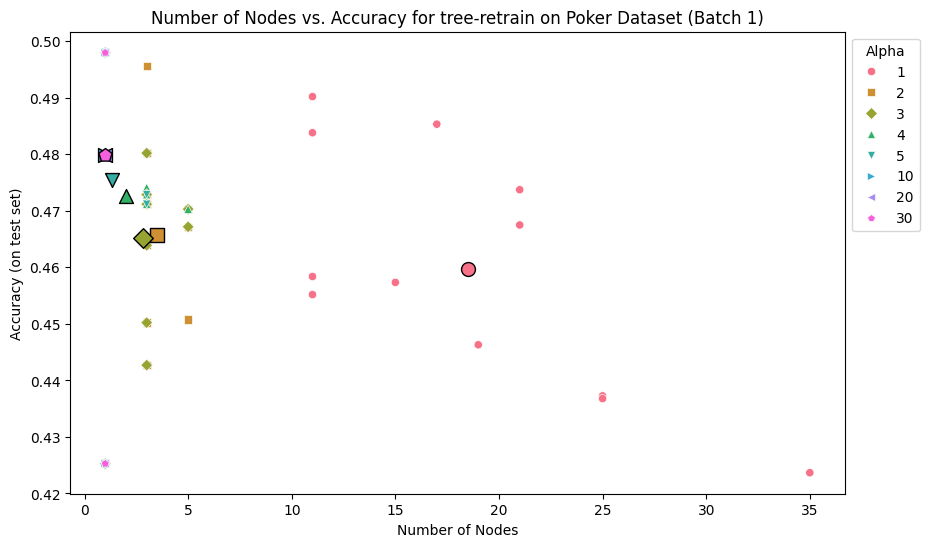}
    \end{center}
    { \hspace*{\fill} \\}
    
    \begin{center}
    \adjustimage{max size={\linewidth}{0.9\paperheight}}{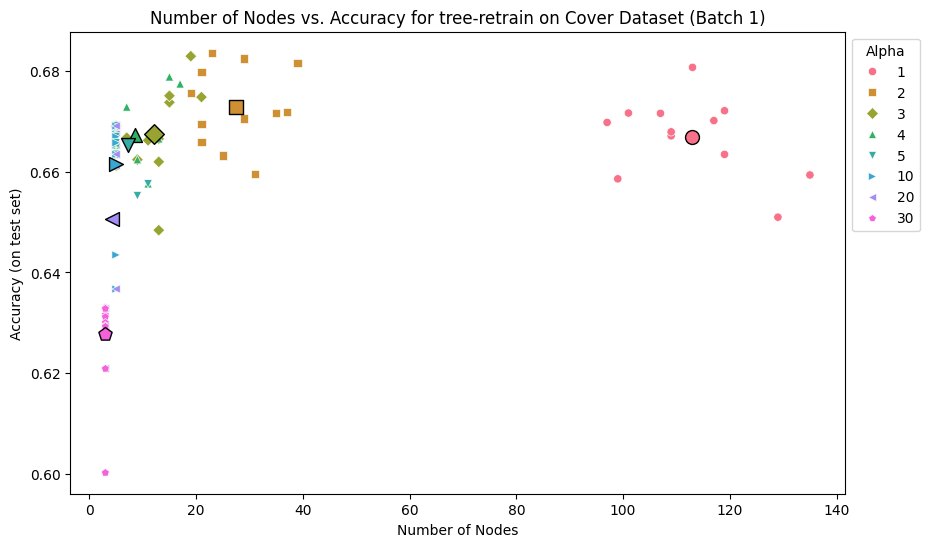}
    \end{center}
    { \hspace*{\fill} \\}
    
    \hypertarget{exploration-of-beta-parameter}{%
\subsection{Exploration of Beta
Parameter}\label{exploration-of-beta-parameter}}

The beta parameter controls the penalty for changed nodes (if a branch
node condition is changed, all nodes under it are also considered
changed). The ability to prioritise a tree that is similar to the one
learned from the previous batch is the novel aspect of the keep-regrow
algorithm.

The visualisations in this section explore the tradeoff between accuracy
(for batch 2, on the test set), number of nodes (for batch 2), and
similarity (between the trees for batch 2 and batch 1) for different
values of beta, at a fixed alpha.

Larger values of beta tend to result in decision trees with less nodes.
This is because adding a new node results in the double penalty of both
additional complexity penalised by alpha and additional changes
penalised by beta.

Larger values of beta result in more similar trees, but comes at the
cost of accuracy. We can see that retraining the tree (shown as
beta=-100 as this is equivalent to not penalising changes) results in
low similarity but high accuracy. Keeping the original treee (shown as
beta=100 as this is equivalent to severely penalising change) results in
100\% similarity, but no improvement in accuracy over the first batch.
Beta allows us to produce trees that fall between these two extremes.

An interesting case is beta=0, which will preserve branches of the
original tree if they don't result in any loss in accuracy on the
training set compared to regrowing those branches. In most cases, this
produces trees with a higher similarity to the original tree than
retraining (beta=-100) but without sacrificing any accuracy.

Based on the results, we suggest a default value of beta=1 to penalise
changes while minimising the loss of accuarcy. On the datasets, we can
see that the drop in accuracy of keep-regrow with beta=1 compared to
retraining a new decision tree (beta=-100) is only a fraction of a
percent.

    \begin{center}
    \adjustimage{max size={\linewidth}{0.9\paperheight}}{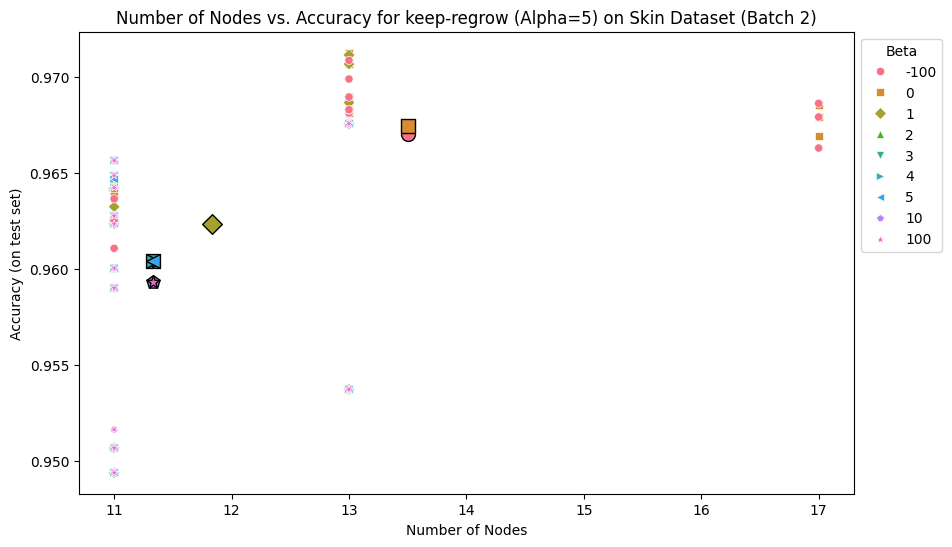}
    \end{center}
    { \hspace*{\fill} \\}
    
    \begin{center}
    \adjustimage{max size={\linewidth}{0.9\paperheight}}{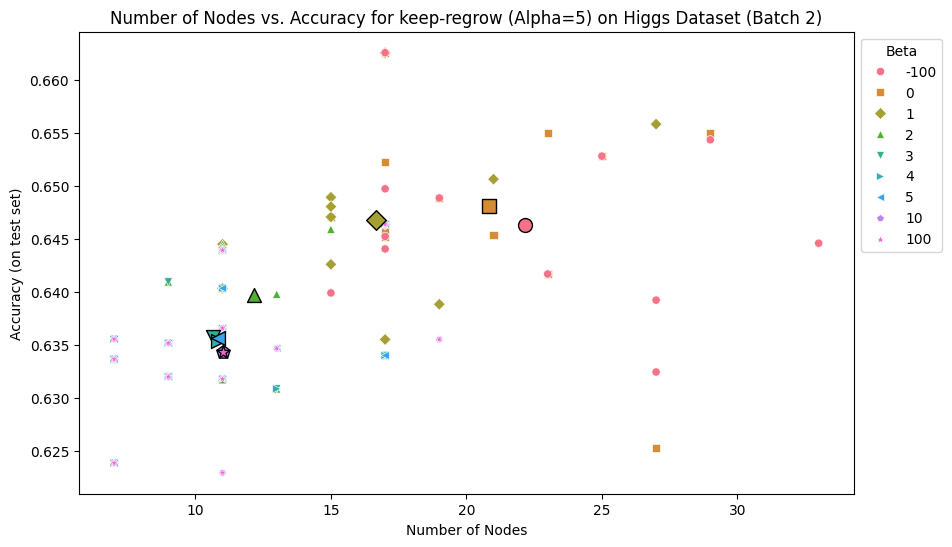}
    \end{center}
    { \hspace*{\fill} \\}
    
    \begin{center}
    \adjustimage{max size={\linewidth}{0.9\paperheight}}{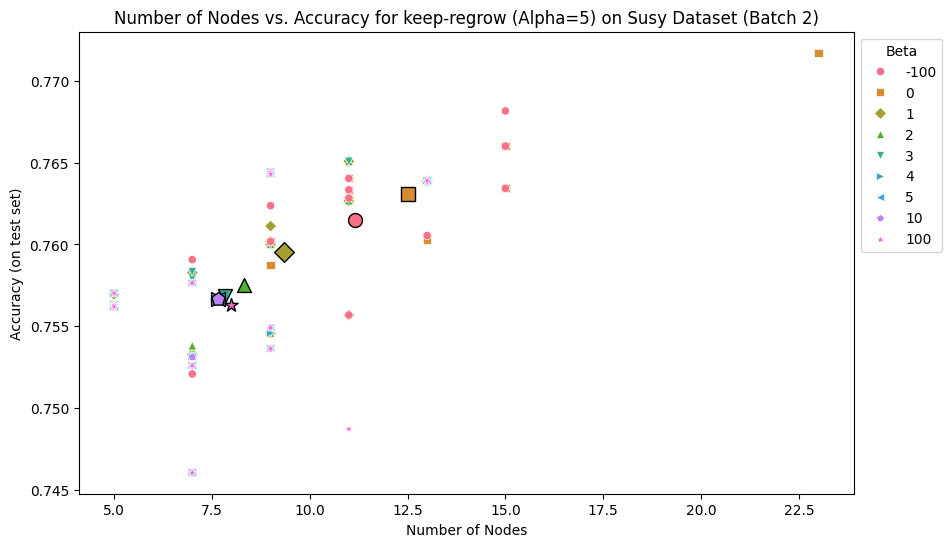}
    \end{center}
    { \hspace*{\fill} \\}
    
    \begin{center}
    \adjustimage{max size={\linewidth}{0.9\paperheight}}{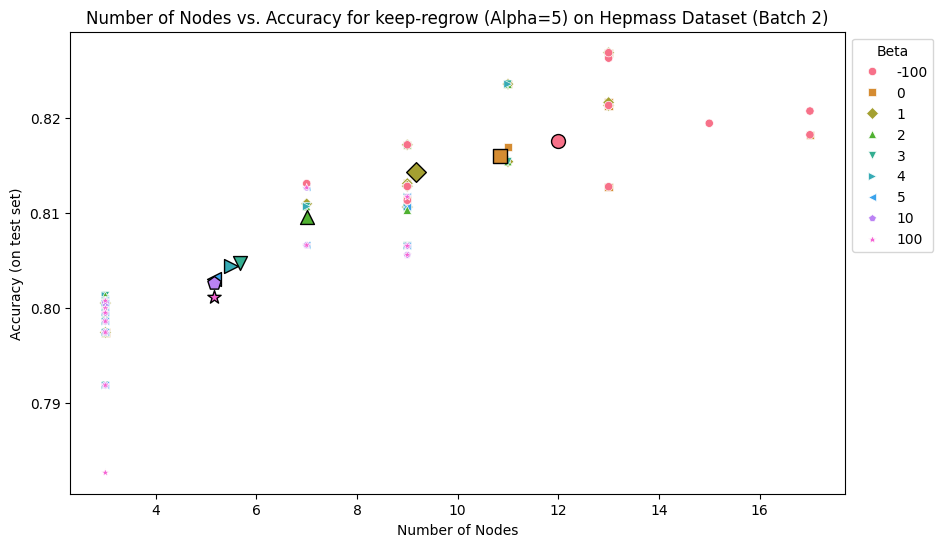}
    \end{center}
    { \hspace*{\fill} \\}
    
    \begin{center}
    \adjustimage{max size={\linewidth}{0.9\paperheight}}{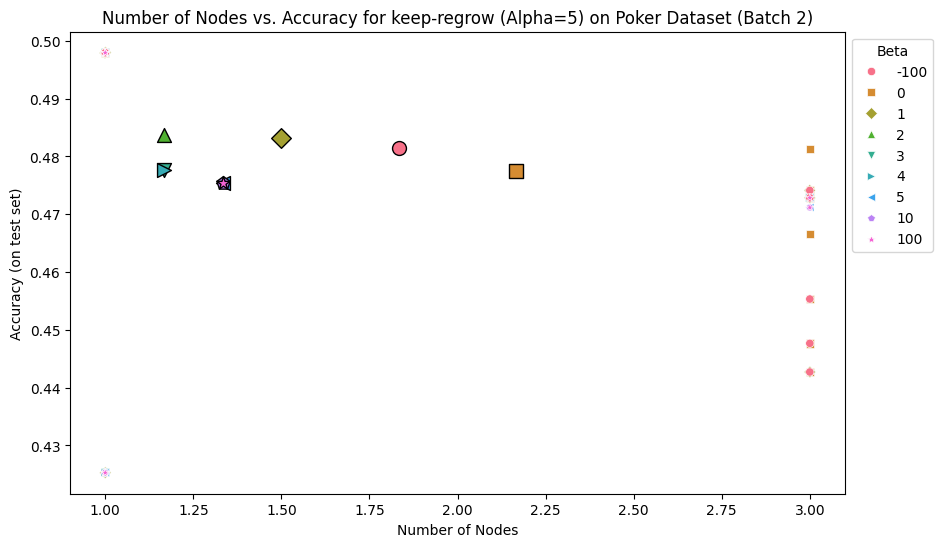}
    \end{center}
    { \hspace*{\fill} \\}
    
    \begin{center}
    \adjustimage{max size={\linewidth}{0.9\paperheight}}{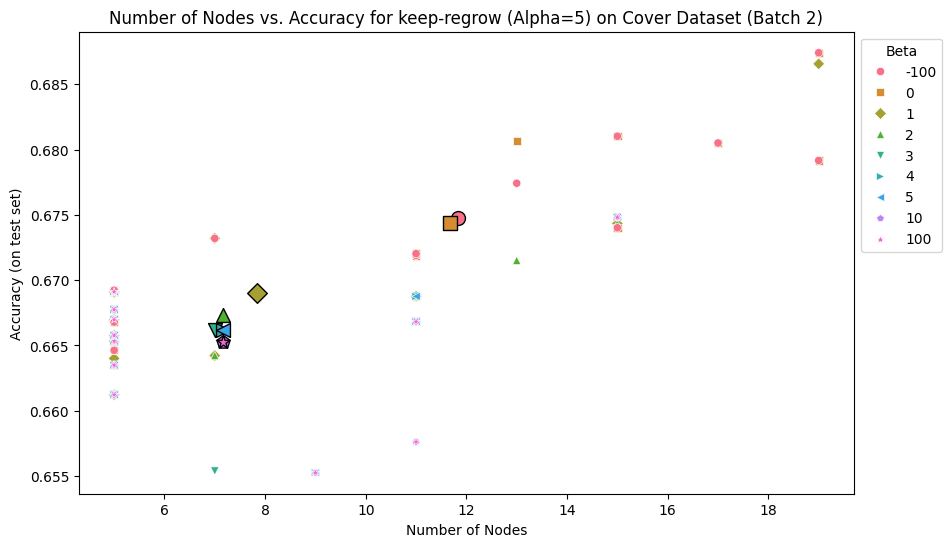}
    \end{center}
    { \hspace*{\fill} \\}

    \begin{center}
    \adjustimage{max size={\linewidth}{0.9\paperheight}}{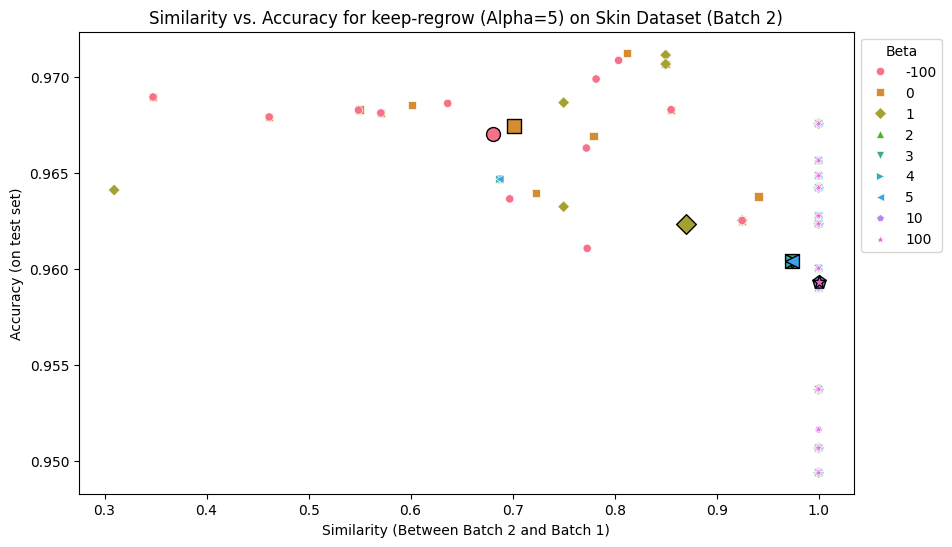}
    \end{center}
    { \hspace*{\fill} \\}
    
    \begin{center}
    \adjustimage{max size={\linewidth}{0.9\paperheight}}{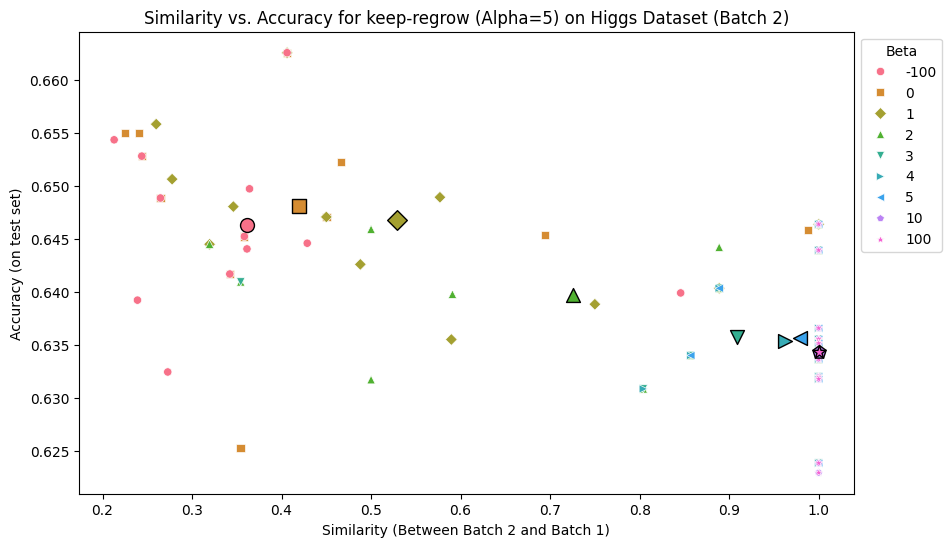}
    \end{center}
    { \hspace*{\fill} \\}
    
    \begin{center}
    \adjustimage{max size={\linewidth}{0.9\paperheight}}{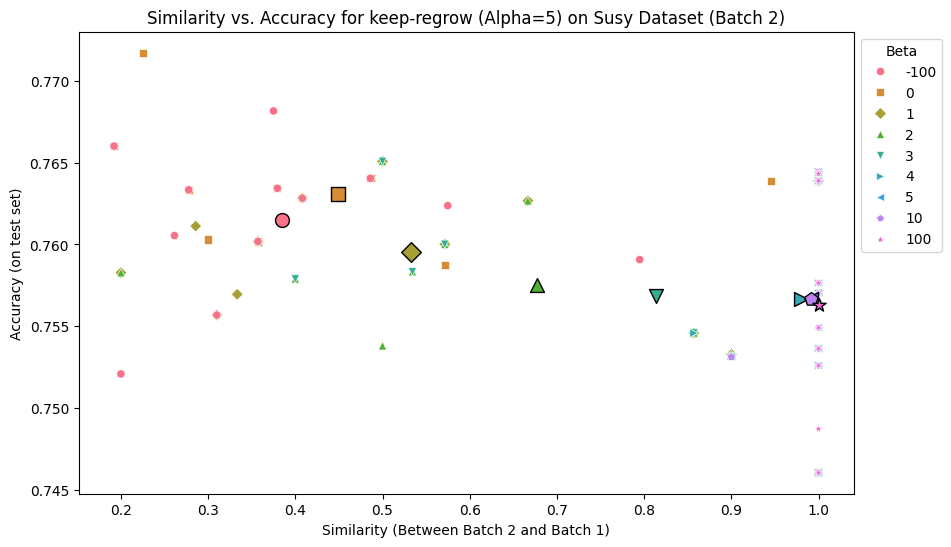}
    \end{center}
    { \hspace*{\fill} \\}
    
    \begin{center}
    \adjustimage{max size={\linewidth}{0.9\paperheight}}{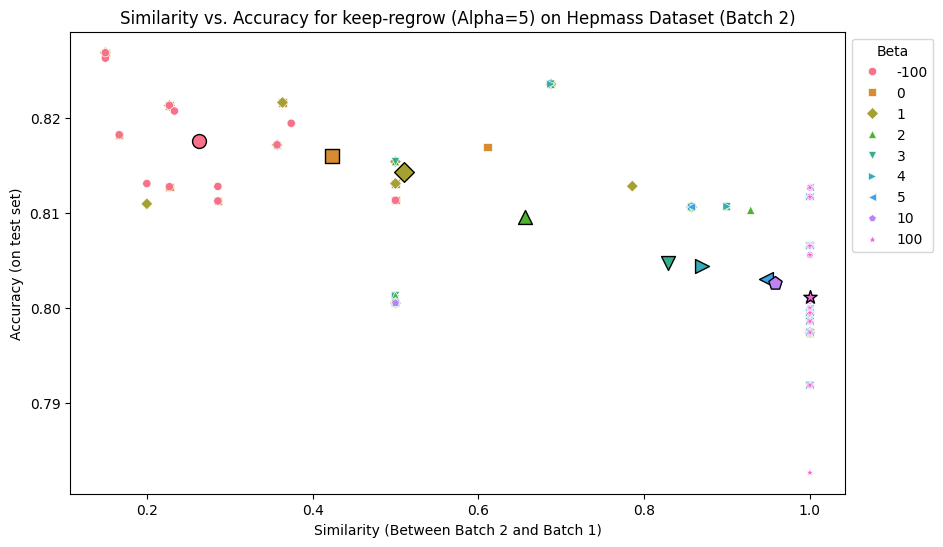}
    \end{center}
    { \hspace*{\fill} \\}
    
    \begin{center}
    \adjustimage{max size={\linewidth}{0.9\paperheight}}{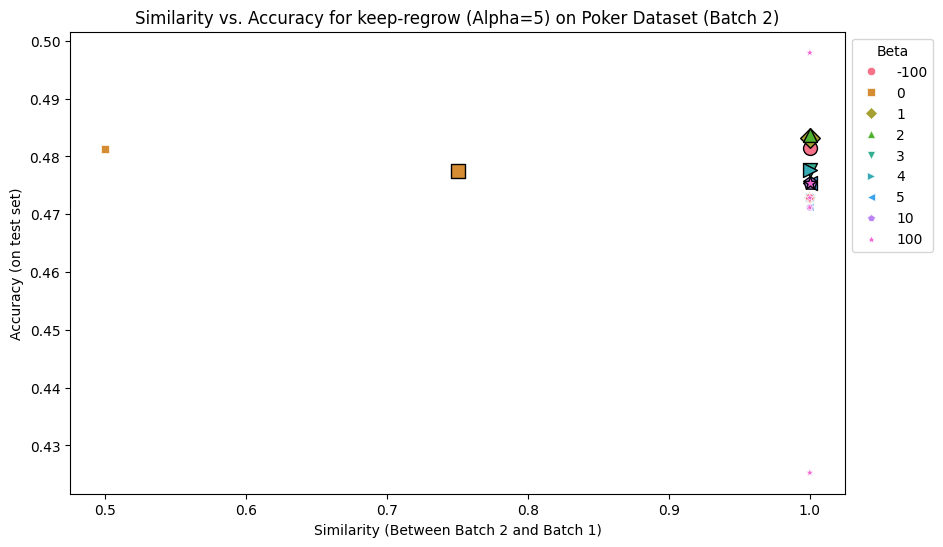}
    \end{center}
    { \hspace*{\fill} \\}
    
    \begin{center}
    \adjustimage{max size={\linewidth}{0.9\paperheight}}{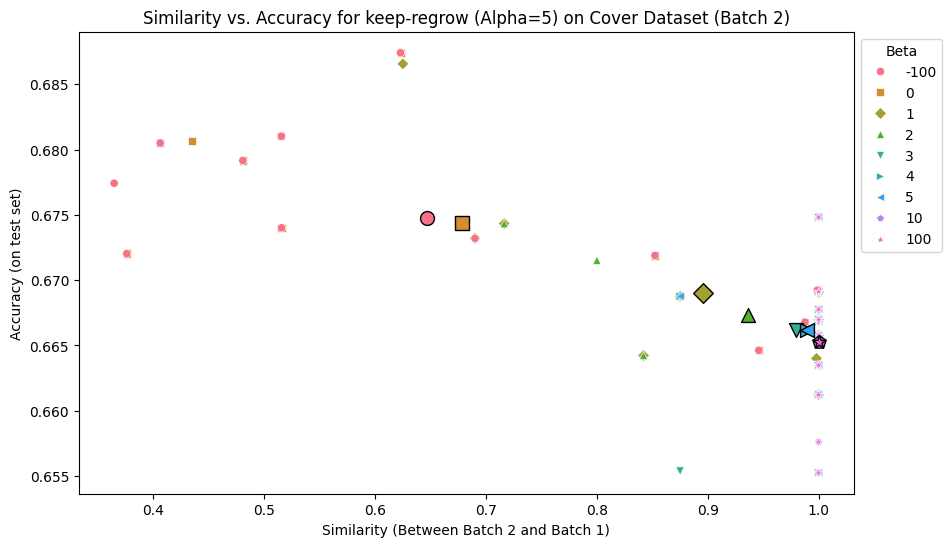}
    \end{center}
    { \hspace*{\fill} \\}
    
    \hypertarget{exploration-of-algorithm}{%
\subsection{Exploration of Algorithm}\label{exploration-of-algorithm}}

The visualisations in this section explore the tradeoff between accuracy
(for batch 2, on the test set), number of nodes (for batch 2), and
similarity (between the trees for batch 2 and batch 1) for different
algorithms.

Keep-regrow falls between re-training (tree-retrain) and keeping the
original decision tree from the first batch (keep-regrow\_5-100) in
terms of number of nodes, similarity, and accuracy.

Compared to EFDT \cite{Manapragada2018} (which is designed for computational efficiency rather
than accuracy), keep-regrow has more nodes and higher accuracy. EFDT
often has a high similarity (due to it's incremental nature), but on
some datasets (Skin, Cover), keep-regrow has a higher similarity than
EFDT

Based on the results, we conclude that keep-regrow offers a unique
option in terms of the trade-off between number of nodes, accuracy, and
similarity.

    \begin{center}
    \adjustimage{max size={\linewidth}{0.9\paperheight}}{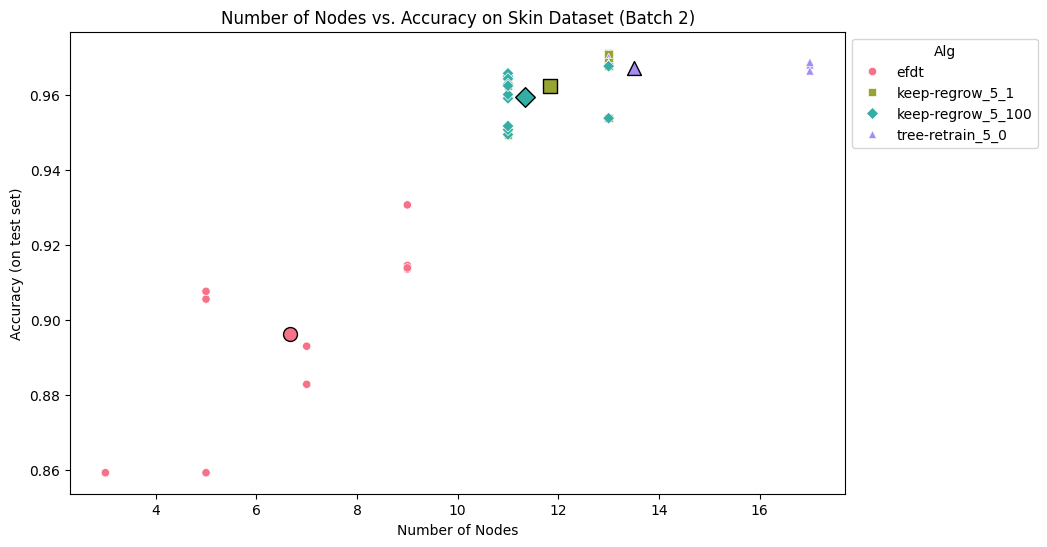}
    \end{center}
    { \hspace*{\fill} \\}
    
    \begin{center}
    \adjustimage{max size={\linewidth}{0.9\paperheight}}{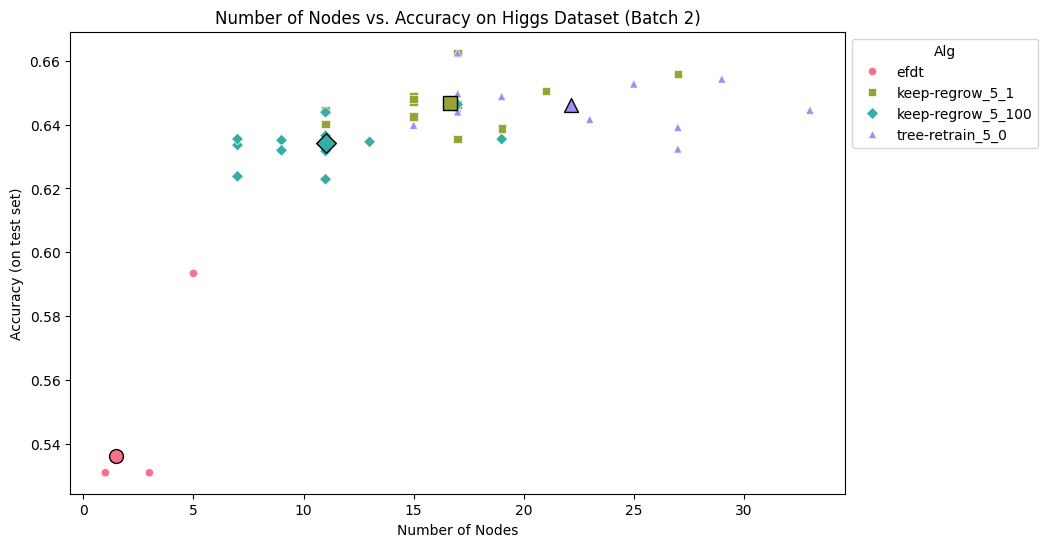}
    \end{center}
    { \hspace*{\fill} \\}
    
    \begin{center}
    \adjustimage{max size={\linewidth}{0.9\paperheight}}{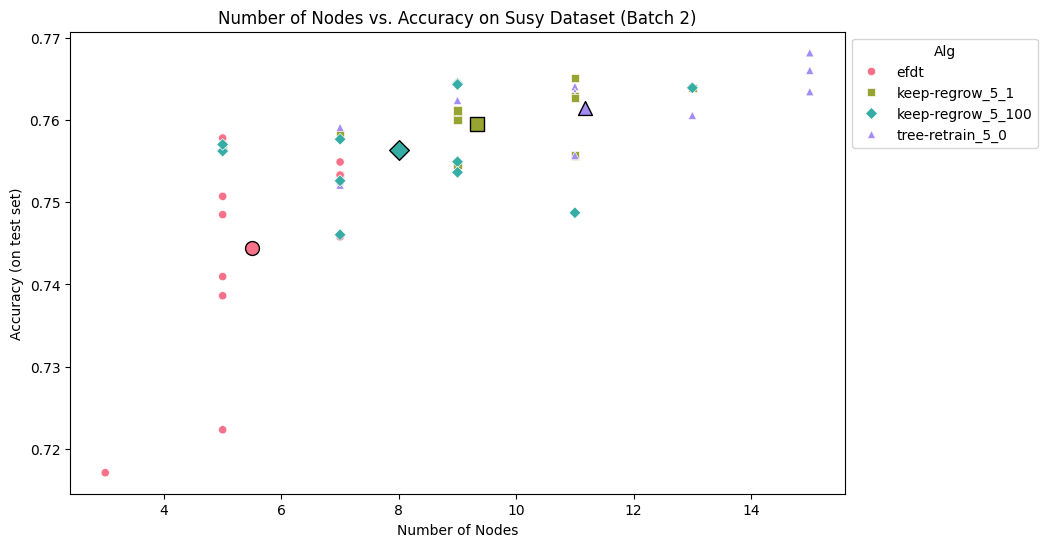}
    \end{center}
    { \hspace*{\fill} \\}
    
    \begin{center}
    \adjustimage{max size={\linewidth}{0.9\paperheight}}{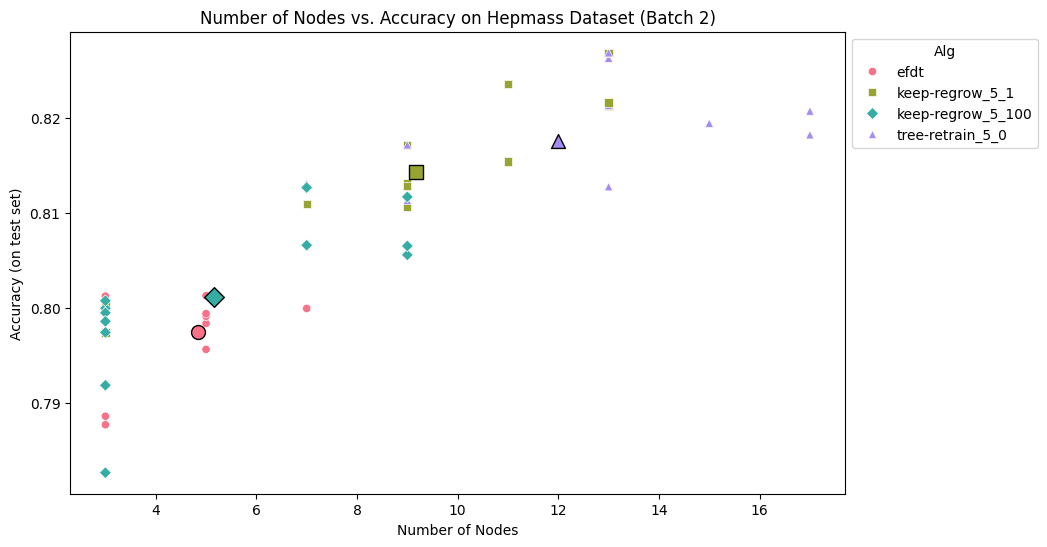}
    \end{center}
    { \hspace*{\fill} \\}
    
    \begin{center}
    \adjustimage{max size={\linewidth}{0.9\paperheight}}{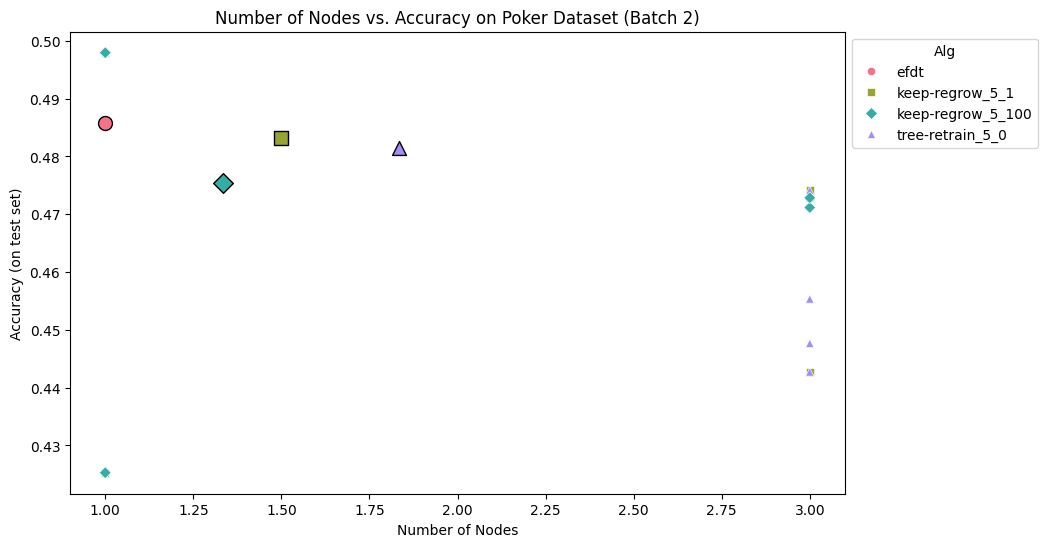}
    \end{center}
    { \hspace*{\fill} \\}
    
    \begin{center}
    \adjustimage{max size={\linewidth}{0.9\paperheight}}{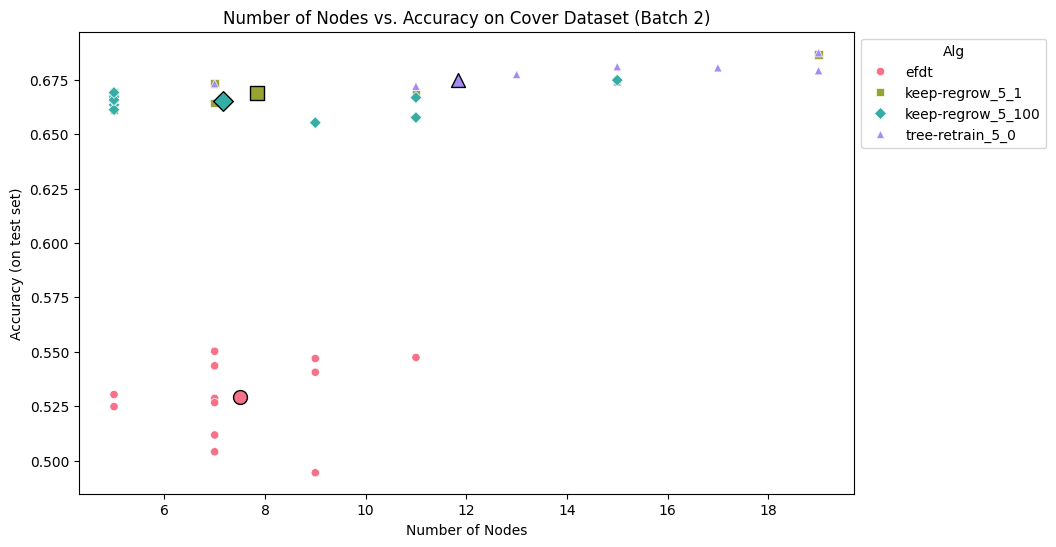}
    \end{center}
    { \hspace*{\fill} \\}

    \begin{center}
    \adjustimage{max size={\linewidth}{0.9\paperheight}}{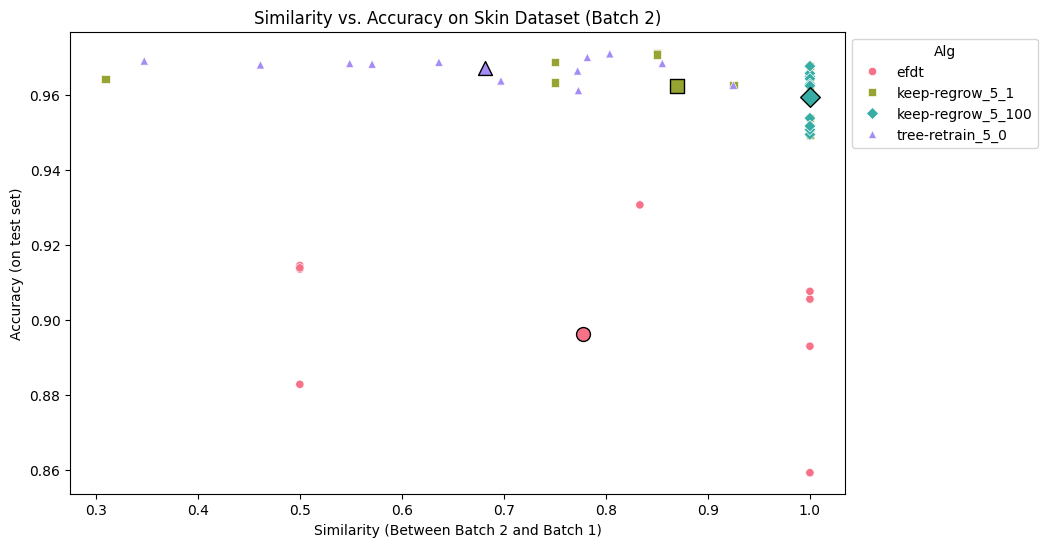}
    \end{center}
    { \hspace*{\fill} \\}
    
    \begin{center}
    \adjustimage{max size={\linewidth}{0.9\paperheight}}{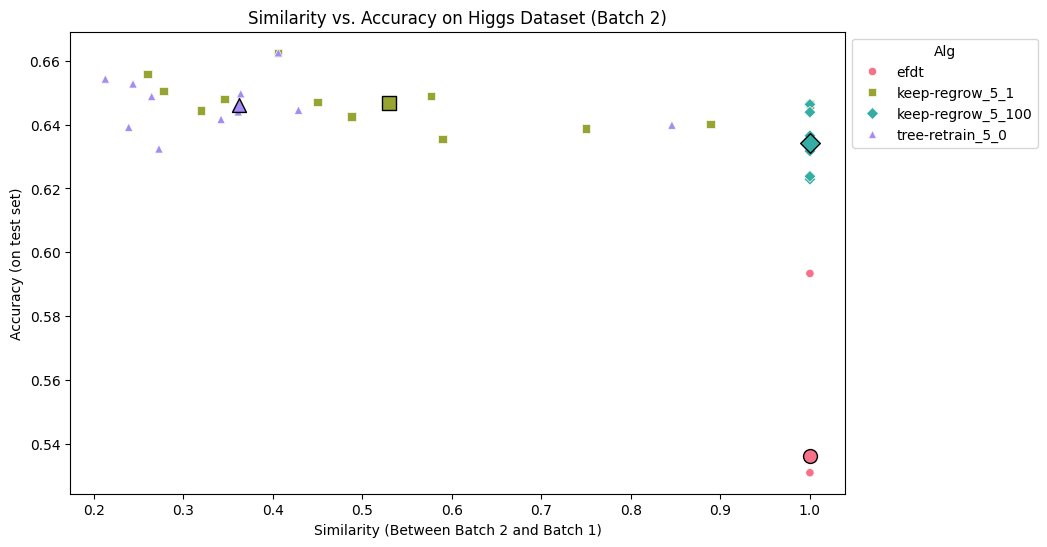}
    \end{center}
    { \hspace*{\fill} \\}
    
    \begin{center}
    \adjustimage{max size={\linewidth}{0.9\paperheight}}{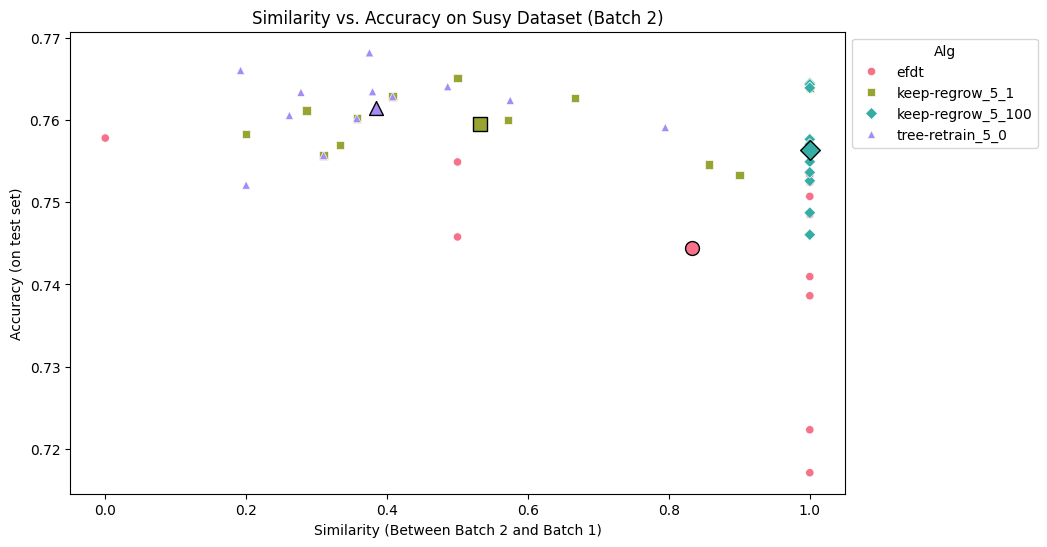}
    \end{center}
    { \hspace*{\fill} \\}
    
    \begin{center}
    \adjustimage{max size={\linewidth}{0.9\paperheight}}{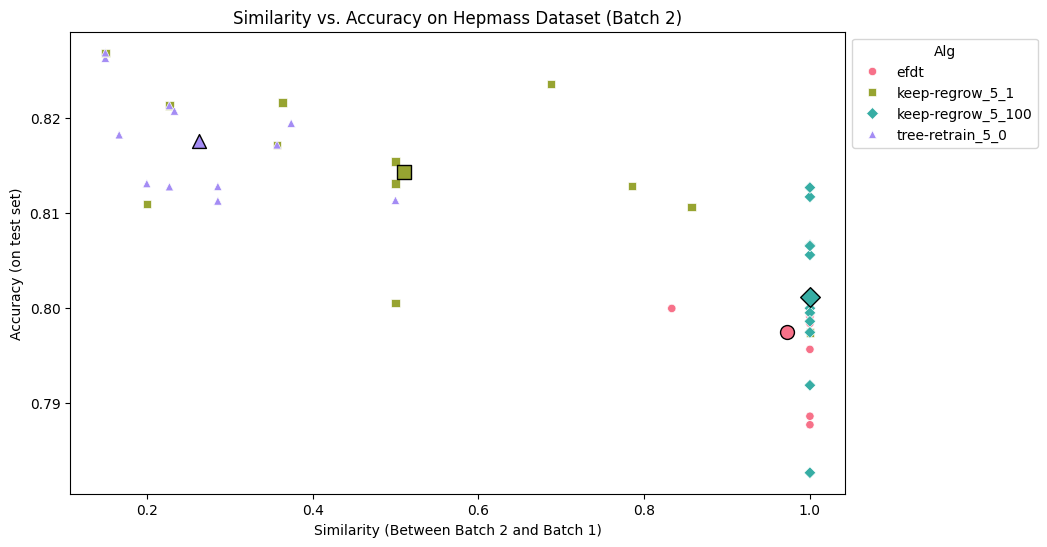}
    \end{center}
    { \hspace*{\fill} \\}
    
    \begin{center}
    \adjustimage{max size={\linewidth}{0.9\paperheight}}{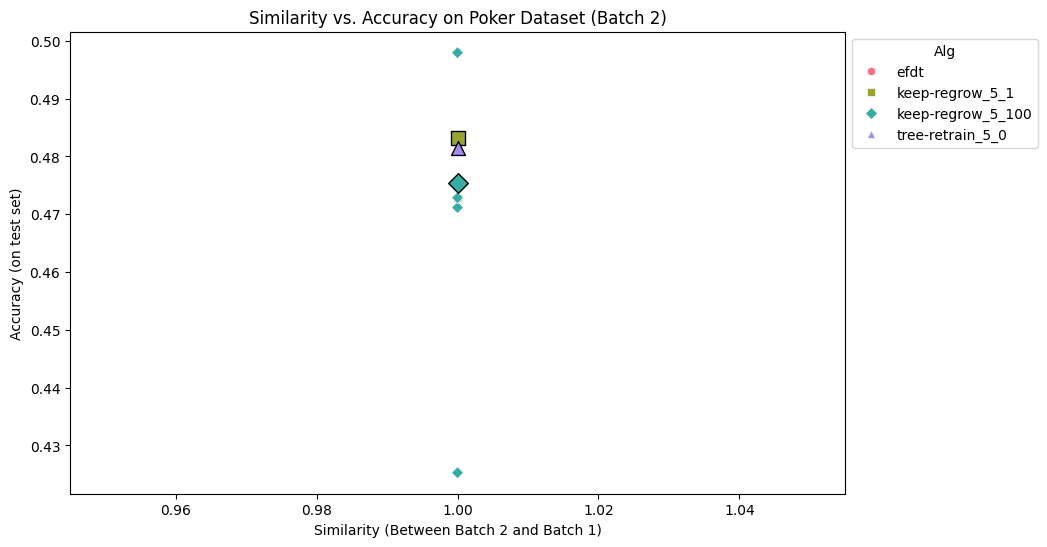}
    \end{center}
    { \hspace*{\fill} \\}
    
    \begin{center}
    \adjustimage{max size={\linewidth}{0.9\paperheight}}{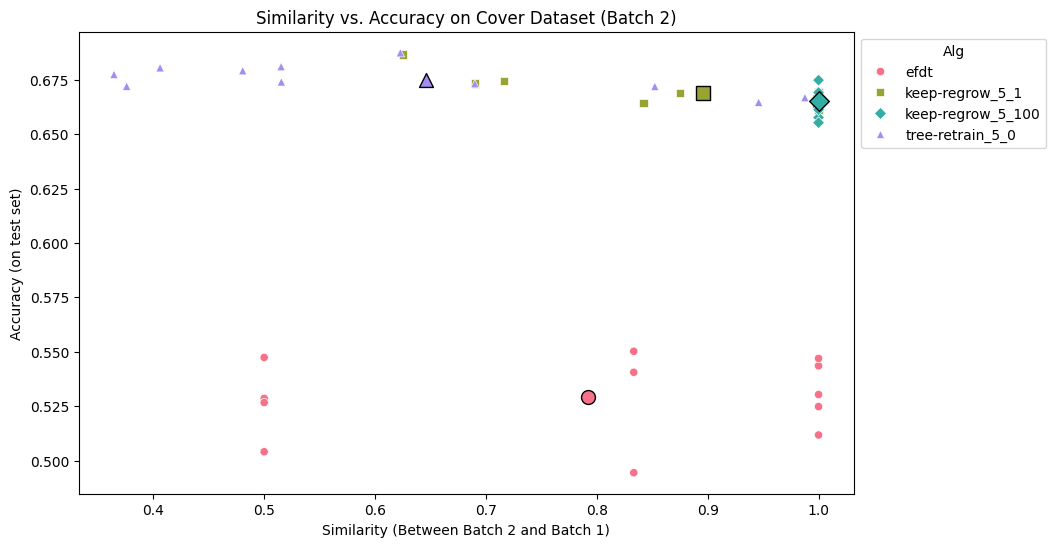}
    \end{center}
    { \hspace*{\fill} \\}
    
    \hypertarget{exploration-of-behaviour-over-time}{%
\subsection{Exploration of behaviour over
time}\label{exploration-of-behaviour-over-time}}

The visualisations in this section explore the accuracy (for batch 1-10,
on the test set), number of nodes (for batch 1-10), and similarity
(between the trees for batch 2vs1, 3vs2, 4vs3, \ldots, 10vs9) for
different algorithms.

We can see that over time (each batch update), keep-regrow is able to
perform similarly to re-training while having a higher similarity.

In the plots of behaviour over time, it is obvious that keeping the
original tree (keep-regrow\_5\_100) is not a good strategy despite 100\%
similarity, as it doesn't allow learning over time.

We can also see that even after 10 batches (1,000 x 10 = 10,000
datapoints), the EFDT algorithm has a lower accuracy than keep-regrow
trained on a single batch (1,000 points). On certain batches EFDT also
has a very low similarity (e.g.~due to changing the root node), whereas
in the experiments, keep-regrow always has a greater than 0 similarity.

    \begin{center}
    \adjustimage{max size={\linewidth}{0.9\paperheight}}{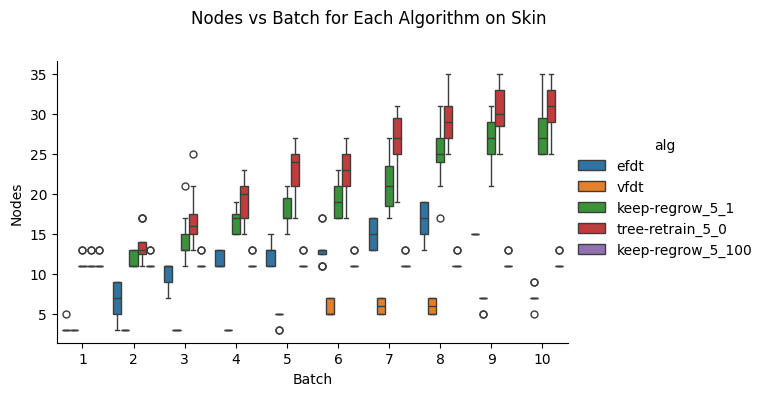}
    \end{center}
    { \hspace*{\fill} \\}
    
    \begin{center}
    \adjustimage{max size={\linewidth}{0.9\paperheight}}{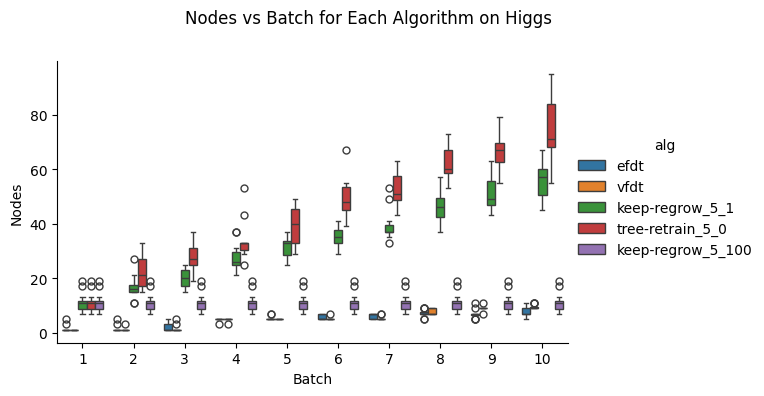}
    \end{center}
    { \hspace*{\fill} \\}
    
    \begin{center}
    \adjustimage{max size={\linewidth}{0.9\paperheight}}{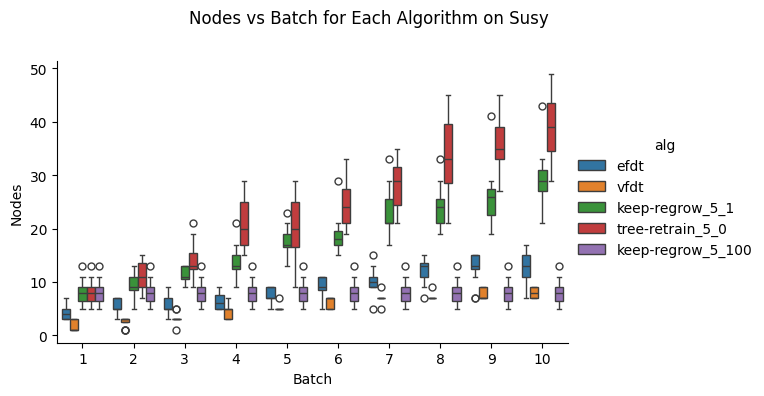}
    \end{center}
    { \hspace*{\fill} \\}
    
    \begin{center}
    \adjustimage{max size={\linewidth}{0.9\paperheight}}{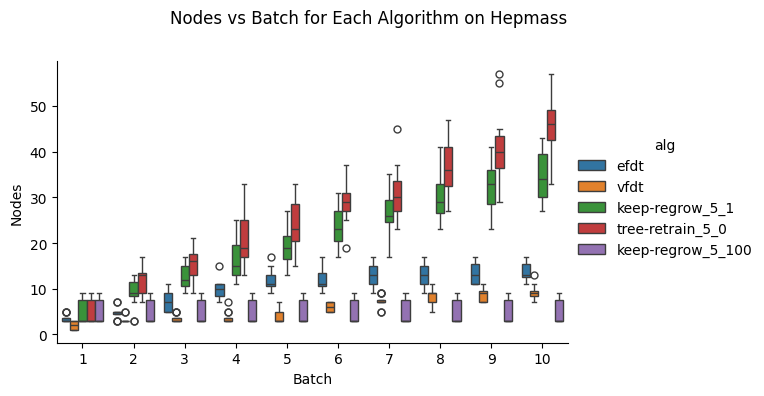}
    \end{center}
    { \hspace*{\fill} \\}
    
    \begin{center}
    \adjustimage{max size={\linewidth}{0.9\paperheight}}{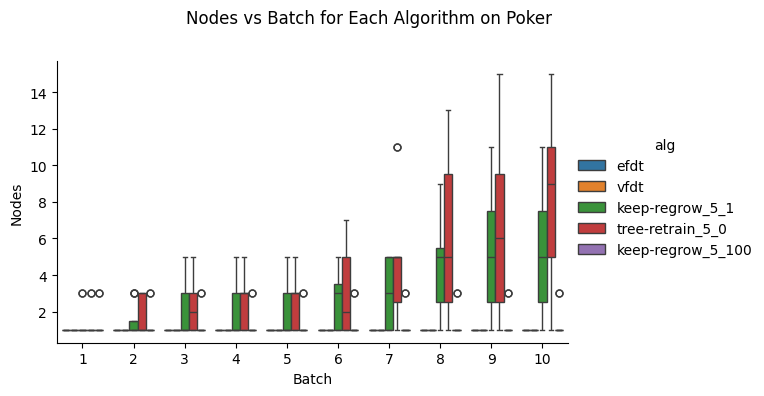}
    \end{center}
    { \hspace*{\fill} \\}
    
    \begin{center}
    \adjustimage{max size={\linewidth}{0.9\paperheight}}{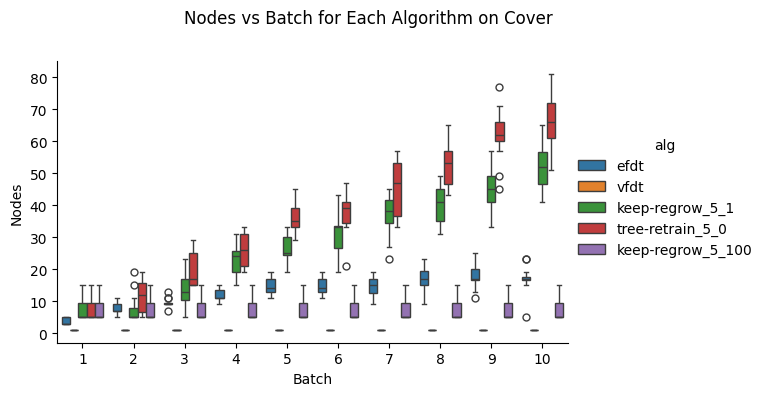}
    \end{center}
    { \hspace*{\fill} \\}

    \begin{center}
    \adjustimage{max size={\linewidth}{0.9\paperheight}}{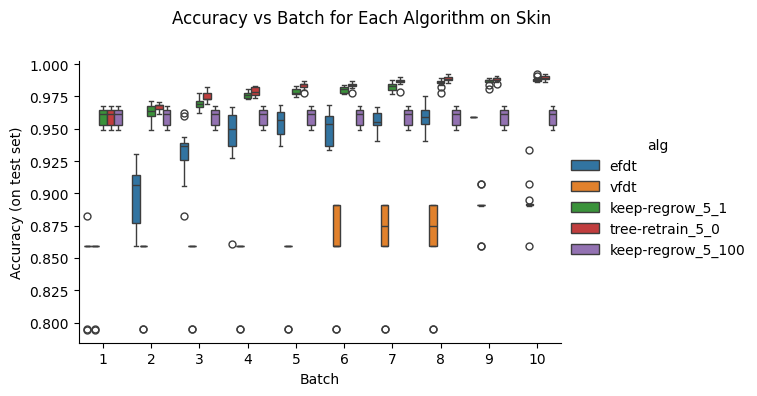}
    \end{center}
    { \hspace*{\fill} \\}
    
    \begin{center}
    \adjustimage{max size={\linewidth}{0.9\paperheight}}{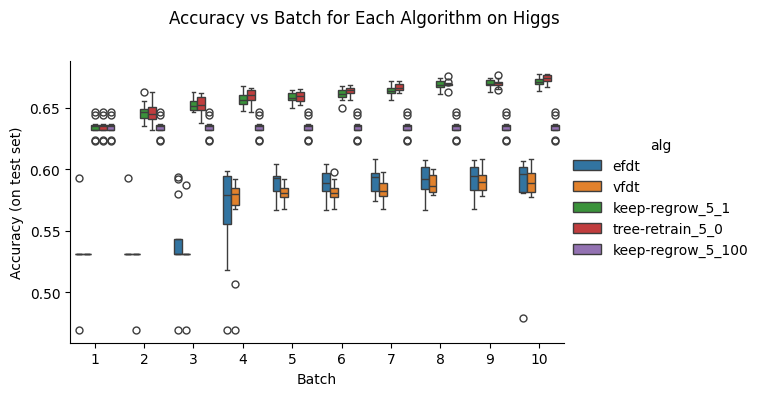}
    \end{center}
    { \hspace*{\fill} \\}
    
    \begin{center}
    \adjustimage{max size={\linewidth}{0.9\paperheight}}{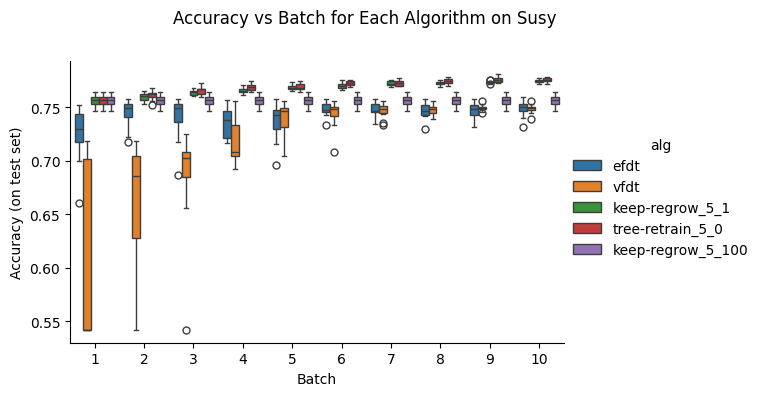}
    \end{center}
    { \hspace*{\fill} \\}
    
    \begin{center}
    \adjustimage{max size={\linewidth}{0.9\paperheight}}{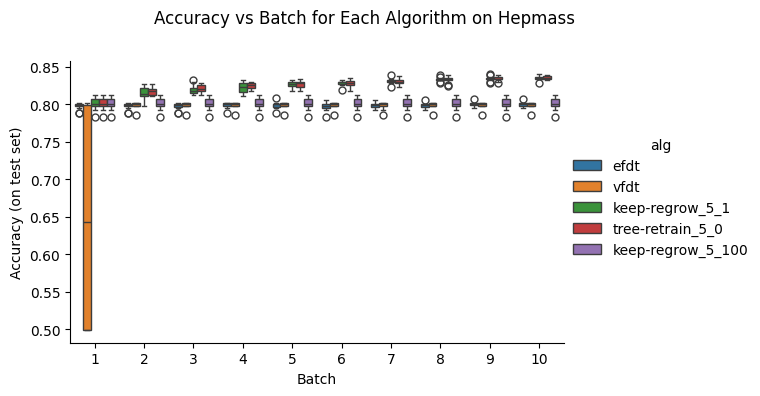}
    \end{center}
    { \hspace*{\fill} \\}
    
    \begin{center}
    \adjustimage{max size={\linewidth}{0.9\paperheight}}{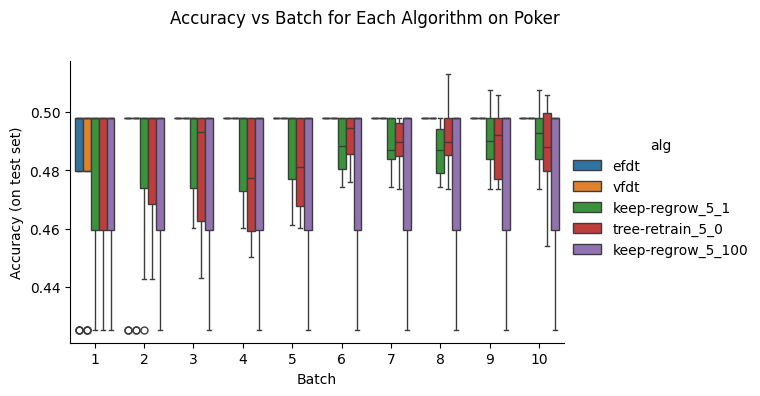}
    \end{center}
    { \hspace*{\fill} \\}
    
    \begin{center}
    \adjustimage{max size={\linewidth}{0.9\paperheight}}{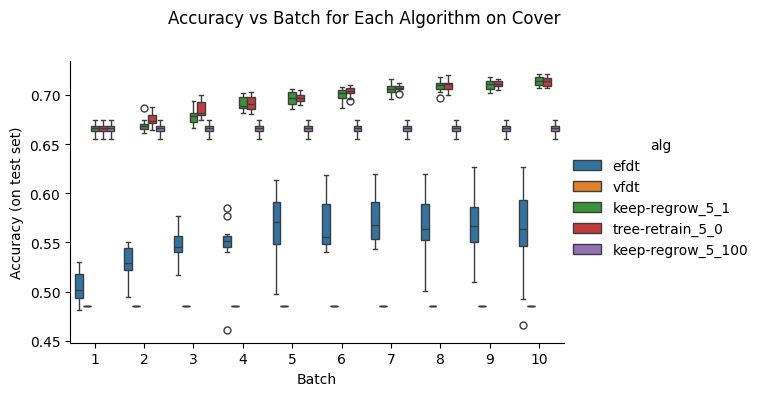}
    \end{center}
    { \hspace*{\fill} \\}

    \begin{center}
    \adjustimage{max size={\linewidth}{0.9\paperheight}}{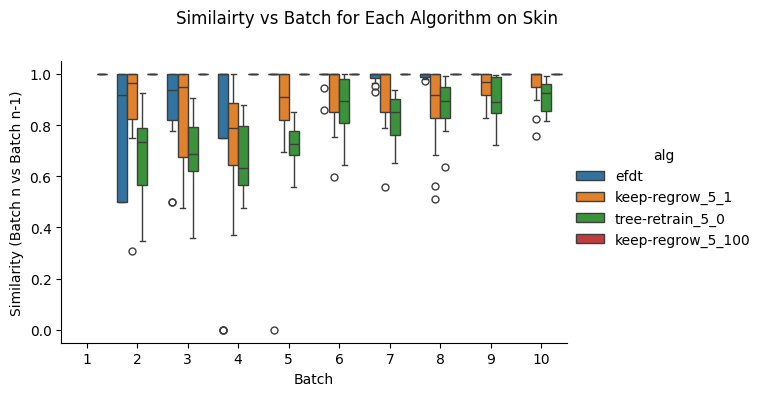}
    \end{center}
    { \hspace*{\fill} \\}
    
    \begin{center}
    \adjustimage{max size={\linewidth}{0.9\paperheight}}{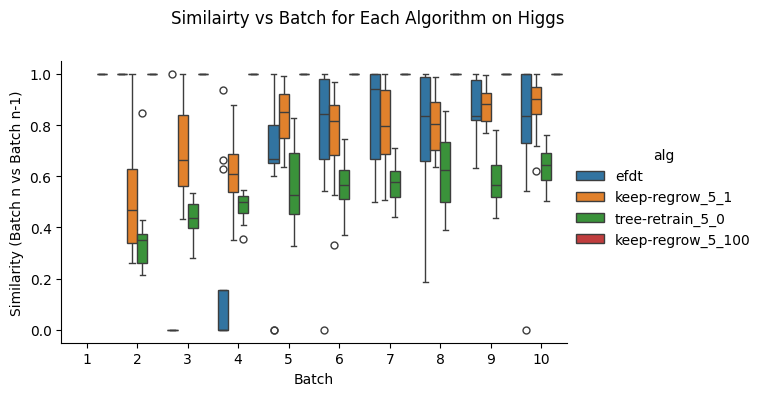}
    \end{center}
    { \hspace*{\fill} \\}
    
    \begin{center}
    \adjustimage{max size={\linewidth}{0.9\paperheight}}{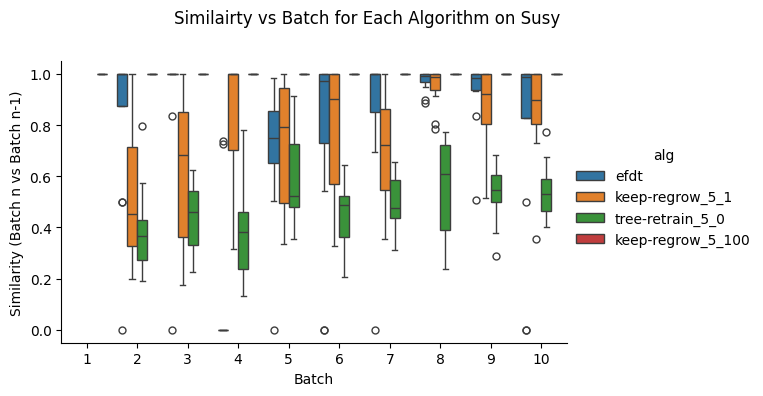}
    \end{center}
    { \hspace*{\fill} \\}
    
    \begin{center}
    \adjustimage{max size={\linewidth}{0.9\paperheight}}{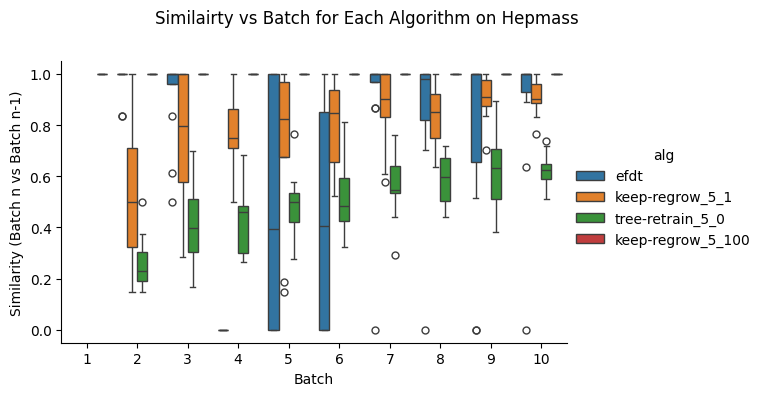}
    \end{center}
    { \hspace*{\fill} \\}
    
    \begin{center}
    \adjustimage{max size={\linewidth}{0.9\paperheight}}{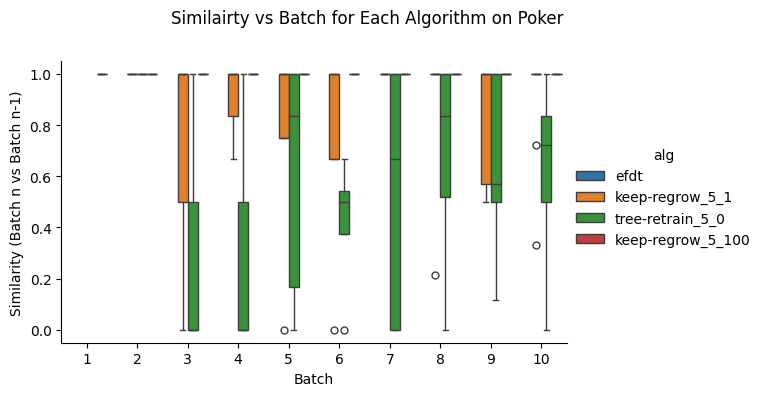}
    \end{center}
    { \hspace*{\fill} \\}
    
    \begin{center}
    \adjustimage{max size={\linewidth}{0.9\paperheight}}{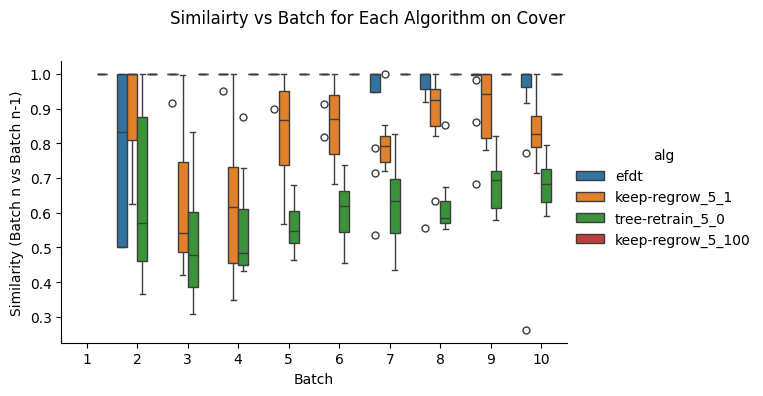}
    \end{center}
    { \hspace*{\fill} \\}


\section{Related Work}


\subsection{Incremental decision trees}

Extremely Fast Decision Trees (EFDT) \cite{Manapragada2018} is an algorithm to incrementally grow a decision tree. The motivation for EFDT differs from ours in that it attempts to minimise changes for computational reasons rather than for human auditing. In contrast, our algorithm is willing to trade off computational efficiency in favour of a decision tree that is easy to audit yet accurate.

In terms of the algorithm, ours only alters a node when the accuracy improvement justifies the cost of a change. In contrast, EFDT will alter a node whenever a new split will result in a statistically significant information gain improvement, regardless of whether this actually improves the overall accuracy. For example, it may just change the ordering of nodes to put the splits with the highest information gain at the top, even though this change to the decision tree doesn't necessarily lead to an improvement in overall accuracy.

\subsection{Similarity search}

Yang et al. \cite{yang2005similarity} propose an efficient approach for estimating the lower bound on the Tree Edit Distance for the purpose of efficient tree similarity search. Such an approach could potentially be incorporated as an alternative for finding similar trees as part of our more general framework for trading off similarity of the updated tree with simplicity and accuracy.

\subsection{Post-hoc explanations}

Tree ensembles are difficult to understand, deploy and debug. The interpretability of tree ensembles is limited due to their complexity \cite{deng2019interpreting}. Deriving an interpretable model from a tree ensemble is referred to as ensemble post-hoc explainability simplification methods. To gain interpretability of tree ensembles, researchers have proposed various methods such as bayesian approaches \cite{hara2018making}, removal of redundant branches and learning sparse weights \cite{yoo2019edit}, and dynamic programming \cite{vidal2020born}. These post-hoc explanation methods of tree ensembles have only been demonstrated on the binary classification \cite{deng2019interpreting, lal2022te2rules, obregon2023rulecosi+, zhao2018iforest}. The experiments on evaluating post-hoc explanation methods are done on smaller datasets \cite{obregon2023rulecosi+}, compared to the study, presented in this paper. Further, inTrees \cite{deng2019interpreting} do not consider rules formed with cross-tree interactions. Similarly, we do not consider cross-tree interactions in our study. Pruning is done as a post-construction method \cite{vidal2020born}, whereas in our study, active pruning is done whilst construction. Also, some of these methods have reported high runtime \cite{lal2022te2rules, obregon2023rulecosi+}. 

Different approaches for making final results of ensemble models interpretable are i) linear combination \cite{benard2021sirus}, ii) representing the model through the generation of new data \cite{vidal2020born}, iii) probabilistic model \cite{wang2017bayesian, hara2018making}, and iv) learning ensemble of base models \cite{deng2019interpreting, mita2020libre}.

Interpretability has been measured in different ways such as i) small number of rules, small number of conditions and small overlap among rules \cite{zhang2020diverse, yang2021learning}, and ii) quality and simplicity of rules \cite{mita2020libre}. However, the diverse definitions make measuring interpretability questionable. Also, interpretability is subjective to humans. Diverse rule sets \cite{zhang2020diverse} are optimised for small overlap between rules. However, the limitations of this algorithm \cite{zhang2020diverse} are the number of conditions in a rule and the parameter to tradeoff between complexity and diversity (i.e. sensitive to parameter setting). 





\section{Conclusions and Future work}

We proposed a framework that incorporates the desire to minimise the number of changes to audit as part of the objective function for growing a decision tree. Our evaluation shows that keep-regrow offers a unique trade-off between number of nodes, accuracy, and similarity. Users have the ability to customise the parameters alpha and beta to control the weighting of different aspects in our framework. Our algorithm generalises other approaches, with always preserving the original decision tree on one extreme where users assign a high weight to minimising changes to audit, to regrowing the entire tree on the other extreme where users assign a low weight to this attribute.

Future work is needed to perform user testing with human participants to test and refine the definitions of complexity and similarity, and to study how these relate to the time taken for humans to audit a decision tree. Further extensions to our framework could assign a higher weight to preserving changes manually made by a human if they reject a decision tree, or more generally, learning from changes suggested by human auditors.

Future work could also look at integrating our framework into the life cycle of real world data collection and analysis projects. At the start of a project, when there is little-to-no data, a human expert or large language model \cite{sivasothy2024large} could generate an initial set of proposed decision tree rules. Our approach could then be used to gradually adapt the decision tree over time as more real world data becomes available.

\bibliographystyle{ACM-Reference-Format}
\bibliography{references}
\end{document}